# Contrastive Learning with Enhanced Abstract Representations using Grouped Loss of Abstract Semantic Supervision

Omri Suissa
Brown University
omri_suissa@brown.edu
omrishsu@gmail.com

Muhiim Ali
Brown University
muhiim_ali@brown.edu

Shengmai Chen
Brown University
shengmai_chen@brown.edu

Yinuo Cai
Brown University
yinuo_cai@brown.edu

Shekhar Pradhan
Brown University
shekhar_pradhan@brown.edu
shprad@gmail.com

## Abstract

*Humans can recognize an image as an instance of a general concept, beyond simply identifying its objects and their relationships. In this paper, we investigate 1. The extent to which VLMs have this concept abstraction capacity, and 2. Strategies for encoding the sort of higher-concept information in images that would enable the resulting VLM model (CLEAR GLASS model) to have this capability to a greater degree. To this end, we introduce a grouped image-caption dataset (MAGIC), which consists of several groups of image captions and for each group a set of associated images and higher-level conceptual labels. We use a novel contrastive loss technique to induce the model to encode in the representation of each image (caption) in a group the information that is common to all members of the image-caption group. Our main contribution is a grouped contrastive loss function based on text-image contrastive groups (outer contrastive loss) as well as an inner loss which measures the distances between image-caption instances in the group. Our training methodology results in the CLEAR GLASS model having the concept abstraction capacity as an emergent capacity because the model is not exposed to the higher-level concepts associated with each group. Instead, the training forces the model to create for each image-caption group a semantic representation that brings it closer to the semantic representation of the higher-level concepts in the latent semantic space. Our experiments show that this training methodology results in a model which shows improvement in abstract concept recognition compared to SOTA models.*

## 1. Introduction

Consider an image featuring a chocolate cake, a table adorned with colorful decorations, and a group of smiling children. A human observer might interpret this scene as a "birthday party," relating it to the broader notion of a "celebration" and ultimately to the even more abstract idea of "social activity." In contrast, current vision-language models (VLMs) [3] tend to focus on recognizing objects [6, 11, 18, 35], their attributes [56] (often termed visual concepts [53]), and, to a limited extent, relationships between objects [1, 52]. However, these models struggle to effectively capture abstract, high-level conceptual meanings in images [14, 23, 50].

Recent efforts to address this limitation have incorporated linguistic supervision [21, 37, 56, 58] to improve relational understanding [5, 17] and extend VLMs beyond object recognition. More recently, HierarCaps [1] explores using linguistic guidance to represent higher-level concepts in image embeddings. However, their training method enforces hierarchical relationships through predefined entailment constraints rather than allowing abstraction to emerge naturally. Furthermore, while HierarCaps[1] provides a hierarchical dataset, its higher-level captions often serve as mere linguistic simplifications of lower-level ones rather than capturing true conceptual abstraction. This limits the model's ability to develop the kind of emergent conceptual understanding that humans possess [4].

To address this gap, we introduce a conceptually hierarchical dataset *(MAGIC)* designed to train models to recognize higher-level aspects of images while incorporating an evaluation framework that assesses both in-distribution and out-of-distribution generalization. Unlike standard evaluations that test models on unseen but similar instances

(*i.e.* in-distribution) [2, 9, 16, 59] or entirely new object categories (*i.e.* out-of-distribution) common in VLM [30, 35, 45], our approach requires the model to generalize across hierarchical levels. Specifically, we train on one level of the hierarchy and test on concepts at a higher level of hierarchy—assessing whether the model can infer relationships that were never explicitly annotated during training. Success in this setting indicates the emergence of true conceptual abstraction.

We further propose a novel grouped caption-image alignment training approach based on the standard image-caption contrastive learning (e.g., CLIP [35]). Instead of aligning individual image-caption pairs, our method aligns groups of captions with groups of images, under three assumptions:

1. The latent space of text embeddings inherently embodies a hierarchical structure of abstraction (Pu et al. [32] share this intuition).
2. Shared conceptual content across a group of thematically closely related captions represents a higher-level concept under which all captions in the group fall.
3. Training on grouped alignments encourages models to infuse the embedding of these higher-level concepts into the image embeddings.

To guide this structured learning, we design a loss function that aligns image embeddings with hierarchical semantic embeddings. Using our dataset, we evaluate whether this approach enables emergent abstraction (*i.e.* the ability to infer higher-level concepts without explicit annotation).

## 2. Related Work

### 2.1. Contrastive Learning in VLMs

With advances in multi-modal embeddings, particularly CLIP [35], contrastive learning on large-scaled web datasets [7, 41, 42] has become the primary approach for training Vision-Language Models (VLMs) [6, 11, 18, 29, 34], enabling superior cross-modal representation learning [3, 5, 38, 48]. As contrastive learning primarily operates at the instance level [26, 49], VLMs like CLIP tend to perform well on fine-grained image-text matching but generalize poorly to novel classes and higher-level concepts [23, 50], unable to capture broader semantic structures or conceptual hierarchies [14]. To improve a model's discriminative power [15, 19, 34], hard negatives are commonly used as contrastive samples. These are samples that are highly similar yet belong to different classes [12, 13, 36, 40]. A standard strategy selects the hardest negatives based on embedding similarity [13, 51], while techniques such as learned embedding-based mining [28, 36], generative hard negatives [22, 38], and feature-based compositional reasoning [55] further enhance contrastive training. *While these techniques improve intra-class distinctions, they are by and large restricted to instance-level comparisons, which render them less suitable for developing semantically richer VLMs that can capture hierarchical relationships.*

### 2.2. GCD and Hierarchical Learning

Generalized Category Discovery (GCD) aims to classify both seen and unseen classes (*i.e.* categories) by using labeled examples of known classes [27, 33], requiring models to infer novel classes without explicit supervision [30, 45]. Existing CLIP-based methods primarily rely on instance-level contrastive learning [31, 32], applying contrastive loss at the sample level while preserving the same hierarchical structure during training and evaluation [47]. Pu et al. [32] propose Dynamic Conceptual Contrastive Learning, which refines contrastive objectives for GCD and defines novelty as classes within the same distribution as seen classes (rather than a higher conceptual level). In contrast, our method introduces novelty abstraction by structuring novelty as a conceptual generalization of seen classes rather than a new class within the same hierarchy. Furthermore, while prior works - including multi-modal GCD [44, 46], guided hierarchical clustering [30], and incremental GCD [57] - optimize for instance-level clustering, our approach focuses on group-based hierarchical discovery. *This lets the model capture shared features that define higher-level novel classes beyond instance-specific learning.*

### 2.3. Abstract Concept Learning

#### 2.3.1. Learning Conceptual Hierarchies

Abstract concept learning in VLMs [56] has largely relied on linguistic supervision [5, 21, 37, 58] and predefined hierarchies [10, 17, 54] to enforce structured generalization. Alper and Averbuch-Elor [1] introduce a contrastive learning approach in which entailment constraints yield hierarchical relationships, with four-tiered positive-negative text-image pairs. Although this method enforces abstraction, it relies on predefined supervision and encodes hierarchical concepts during training directly. Other works explore abstraction differently: Hernández-Cámara et al. [17] assess human-CLIP alignment across abstraction levels but do not model hierarchical emergence. Roth et al. [37] and Zheng et al. [58] show that textual embeddings improve visual classification by introducing high-level descriptors that focus on semantic enrichment (rather than structural emergence). *Our approach diverges by enabling conceptual hierarchies to emerge naturally from grouped image-caption associations (i.e. during training, the model does not see the abstract concepts at all), adapting contrastive learning with hard negatives for a more robust and generalized abstraction learning paradigm.*

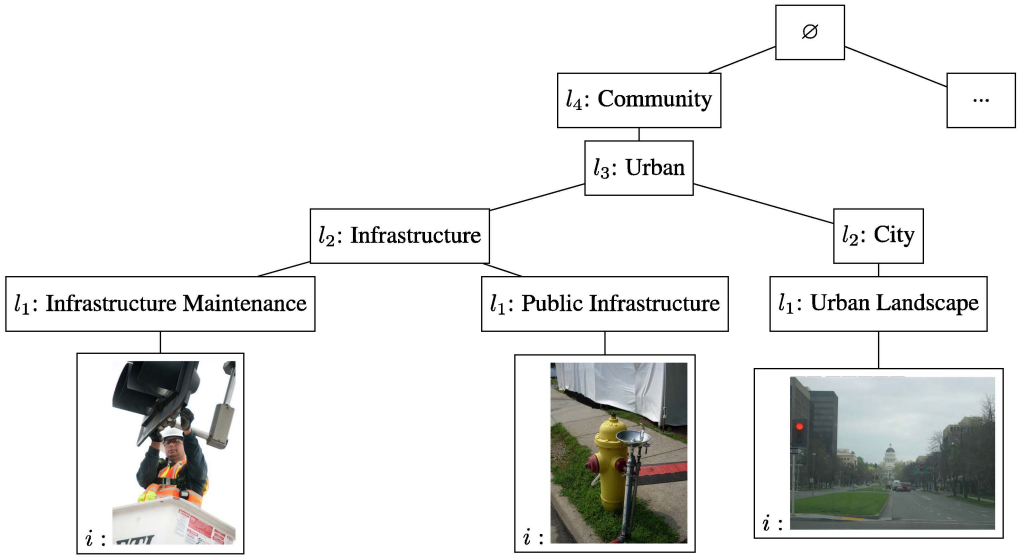

Figure 1. In the *MAGIC* dataset, a visual-semantic hierarchy organizes each image by linking it to an abstract concept node derived from its caption. These nodes form a hierarchical tree in which higher levels represent increasingly abstract concepts. Each level (*i.e. l*) is noted with its distance from the image-caption pair (*i.e. i*).

### 2.3.2. Abstract Concept Benchmarks

Existing benchmarks for abstract concept learning primarily rely on syntactic simplification or categorical generalization rather than true emergent abstraction. HierarCaps [1] constructs hierarchical captions by both textual and lexical entailment [20], creating hierarchies through text simplification rather than true conceptual emergence. Similarly, BREEDS [39] provides structured subclass generalization tasks but was not designed for abstract concept inference. While Object Concept Learning (OCL) [24] captures conceptual similarity beyond object recognition, it does not model hierarchical abstraction. *Our dataset, MAGIC, instead structures abstraction concepts through a graph-based representation, enabling models to infer hierarchy dynamically rather than from predefined text constraints.*

## 3. Method

To address the challenge of inferring highly abstract concepts from images via linguistic supervision, we introduce both a novel dataset and a new loss function. We hypothesize that the centroid (*i.e.* the average embedding) of image-caption joint embedding sharing the same abstract concept approximates the underlying abstract concept in the embedding space. Based on this assumption, we train the model to distinguish between groups of images and, at the same time, to pull each group's embeddings toward their centroid, thereby encouraging the emergence of abstract concept inference (see Sec. 3.2). This is in contrast to the common practice in contrastive learning of training directly on image-caption pairs, which might not capture true abstraction. Notably, this capacity of abstract concept understanding is emergent rather than explicitly learned, as the model never encounters explicit abstract concept labels during training. In this respect, our task shares similarities with the Generalized Category Discovery (GCD) task [45], in which abstract concepts act as novel classes. However, unlike GCD, our abstract concepts are hierarchical, and the model is never directly exposed to any of them during training. This approach aims to make the model more robust and usable in real-world scenarios.

As illustrated in Fig. 1 and detailed in Sec. 3.1, our approach begins with constructing the Multimodal Abstraction of Grouped Images and Concepts (*MAGIC*) dataset. This dataset enables the model to enhance its representation of abstract concepts in the joint embedding space during training. Additionally, we propose Contrastive Learning with Enhanced Abstract Representations using Grouped Loss of Abstract Semantic Supervision (*CLEAR GLASS*) model (Sec. 3.2), a variant of the CLIP [35] model trained with a novel loss function that fosters the emergence of abstract concept inference within a contrastive learning framework. As shown in our experiments (Sec. 4), this approach improves VLM's understanding of abstract concepts.

### 3.1. MAGIC Dataset Generation

Although HierarCaps [1] creates hierarchical datasets, it lacks the linguistic control and level of abstraction required for our approach. In many cases, higher-level concepts are represented in HierarCaps by simplified or truncated captions (*e.g.* HierarCap's first-level abstraction prompt instructs to "Shorten the following image caption," which does not capture true abstraction by our definition). Therefore, we constructed a dataset that genuinely reflects the linguistic abstraction present in captions.

To represent abstract concepts, we construct a hierarchical tree of concepts in which images (and their corresponding captions) are organized into multiple levels of semantic abstraction. This structure allows us to evaluate the extent to which the model can represent abstract concepts beyond the common simple image-caption alignment in other contrastive learning datasets.

As detailed in Algorithm 1, the dataset generation process comprises the following steps:
1. Infer an initial abstract concept for each caption.
2. Derive multi-level abstract concepts.
3. Construct a directed acyclic graph (DAG) based on the abstraction levels.
4. Merge small nodes based on semantic similarity.
5. Generate a generalized caption for each image based on the concept of its node.
6. Mine hard negative concepts for each node.

**Algorithm 1** MAGIC Dataset Generation

**Require:** $l_{max}, Size_{min}, Sim_{min}$
1: $DAG_{abs} \leftarrow DAG()$
2: **for** $c_i \in COCO$ **do** ▷ Sec. 3.1.1
3:     $abs_{child} \leftarrow LLM_{abs}(c_i)$
4:     $abs_{child}.append(c_i)$
5:     **for** $level < l_{max} - 1;\ level\text{++}$ **do**
6:         $abs_{level} \leftarrow LLM_{abs}(abs_{child})$
7:         $abs_{level}.append(abs_{child})$
8:         $abs_{child} \leftarrow abs_{level}$
9:     **end for**
10:     $DAG_{abs}.append(abs_{child})$
11: **end for**
12: **for** $n \in DAG_{abs}.level(l_{max} - 1)$ **do** ▷ Sec. 3.1.2
13:     **if** $n.children < Size_{min}$ **then**
14:         $Sim_n \leftarrow DAG_{abs}.sim(n)$
15:         **if** $\cos(Sim_{node}, n) \geq Sim_{min}$ **then**
16:             $DAG_{abs}.merge(n, Sim_{node})$
17:         **else**
18:             $DAG_{abs}.remove(n)$
19:         **end if**
20:     **end if**
21: **end for**
22: **for** $n \in DAG_{abs}.level(l_{max})$ **do** ▷ Sec. 3.1.3
23:     $n.append(LLM_{gen}(n))$
24: **end for**
25: **for** $n \in DAG_{abs}.level(l_{max} - 1)$ **do** ▷ Sec. 3.1.4
26:     $n.negs \leftarrow DAG_{abs}.sim\_many(n, Sim_{min})$
27: **end for**
28: **return** $DAG_{abs}$

where $l_{max}$ denotes the maximum abstraction depth to generate, $Size_{min}$ specifies the minimum number of images per node (*i.e.* group), and $Sim_{min}$ defines the minimum similarity threshold for merging small nodes. $LLM_{abs}$ is a function that infers a general concept from a caption (or concept) using a large language model. The function $DAG.level$ returns all nodes at a specific level, $DAG.sim$ identifies the most similar node based on cosine similarity ($\cos = \frac{x \cdot y}{|x||y|}$), and $DAG.sim\_many$ retrieves all nodes with similarity above the threshold. Finally, $DAG.merge$ merges the children of two nodes (merging the smaller node into the larger one), and $LLM_{gen}$ generates a generalized caption based on a node and its parent concepts.

### 3.1.1. Caption Processing & Hierarchical Inference

To evaluate the visual-language model's (VLM) capacity to infer abstract concepts, we begin by processing the raw captions associated with each image in the COCO dataset to extract their first-level ($l_1$) semantic concept. For each image (which may have up to five captions), a large language model (LLM) is employed to generate an initial, generalized category that encapsulates the image's primary semantic concept (see supplementary materials for prompts and details). These initial categories serve not only as the first-level abstraction labeling, but also provide the basis for further abstraction inference. Based on these labels, we derive three additional levels ($l_1$ - $l_4$) of abstraction using an LLM. At each subsequent level, the parent (*i.e.* superclass) of the current concept is retrieved, thereby constructing a multi-layered hierarchy that progresses from specific descriptions to highly abstract concepts.

### 3.1.2. Graph Construction

Utilizing the multi-level abstractions generated in the previous stage, we construct a directed acyclic graph (DAG) where each node represents a distinct concept at a specific level of abstraction. In this graph, edges denote the parent-child relationships. Each node aggregates images and their captions that share the corresponding concept. Given that some nodes may contain only a small number of images, a pruning step is introduced to ensure robust group formation. Specifically, nodes with fewer than five images are merged with linguistically similar nodes based on cosine similarity between their concept embeddings. If the similarity exceeds a threshold ($\cos > 0.9$), the smaller node is merged into the larger one; if no such similar node exists, the small node is removed from the graph. This process guarantees that the final DAG comprises robust, semantically coherent clusters that are optimally structured for group-based contrastive learning.

### 3.1.3. Caption Generalization

To promote group-based learning and enforce semantic closeness within each node, we generate generalized captions that reflect the common semantic information of all captions within a group. A large language model is utilized to convert these specific captions into a generalized variant, ensuring that the resulting captions retain the core abstract concepts information (see supplementary materials for more details). By doing so, the model is encouraged to learn representations that are invariant to individual caption variations and are instead focused on the shared, higher-level semantics of the group.

### 3.1.4. Hard Negative Mining

The selection of negative examples is pivotal in contrastive learning, as it directly influences the quality of the learned representations [52]. In our approach, we extend the concept of hard negative mining to the abstract concept level rather than operating solely at the individual caption or image level. Specifically, for each node in the DAG, we identify hard negatives by computing the cosine similarity between concept embeddings. If the semantic similarity between two distinct concepts exceeds a set threshold ($\cos > 0.85$), the concept is treated as a hard negative with respect to the other. These hard negatives are particularly challenging because they lie close to the decision boundary, which requires the model to learn finer distinctions between similar abstract concepts. Integrating hard negative mining strategy into our dataset generation pipeline not only enhances the discriminative power of the model but also bolsters its ability to differentiate between subtle variations in abstract concepts. The training and evaluation datasets of CLEAR GLASS model (Sec. 3.2) consist entirely of the hard negatives of each group, forcing the model to attend to the nuanced differences between the hard negatives.

| Measure | *MAGIC* dataset |
|---|---|
| Total image-caption pairs | $81,602$ |
| Level 1 ($l_1$) nodes | $1,071$ |
| Level 2 ($l_2$) nodes | $281$ |
| Level 3 ($l_3$) nodes | $113$ |
| Level 4 ($l_4$) nodes | $87$ |
| Avg. images per node | $32.48$ |

Table 1. *MAGIC* dataset statistics

## 3.2. CLEAR GLASS Model

The core intuition behind the *CLEAR GLASS* model is that the latent space of text embeddings inherently embodies a hierarchical structure of abstraction ([32] share this intuition and validate it in the context of GDC). In this framework, a high-level concept such as *Social Gathering* is positioned near the centroid (*i.e.* middle) of its constituent lower-level concepts, for instance, *Celebration* and *Conferences*. Similarly, *Celebration* lies near the centroid of more specific concepts like *Birthday* and *Wedding*. Moreover, the embeddings of sentences in a collection (*e.g.* "Mike went to his friend's birthday party", "Jane and I are getting married") converge toward the centroid (*i.e.* the common representation) of their shared higher-level concept (*e.g.* Celebration). Based on this intuition, we employ linguistic supervision via contrastive learning to transfer semantic features to image embeddings, thereby emphasizing the abstract concept in the joint latent space (*i.e.* image and caption). We refer to this method as Group Contrastive Learning because it encourages the model to focus on shared features within the group rather than on individual pairs.

### 3.2.1. Group Contrastive Learning

Building on this intuition, as shown in Fig. 2, we propose a loss function that will lead to a dual effect:

1. Encourage the model to differentiate between distinct groups of captions and images.
2. Enhance the internal semantic tightness of each group (*i.e.* draw samples closer to their respective centroids).

We refer to the first component as the *outer group loss* and the second as the *inner group loss*. To implement these ideas, we introduce two variants of contrastive learning: Pairwise Oriented Contrastive Learning and Centroid Oriented Contrastive Learning.

### 3.2.2. Pairwise Oriented Contrastive Learning

The idea behind the Pairwise Oriented Contrastive Learning is not only to force each individual image and text to match within a group (as common in contrastive learning) but also to ensure that the joint representation (*i.e.* the interaction between image and caption) is semantically tight across the group. Formally, we define it as follows: For each group $g = 1, \ldots, M$ we have $N$ image–text pairs $\{(I_{g,i}, T_{g,i})\}_{i=1}^{N}$ with $I_{g,i}, T_{g,i} \in \mathbb{R}^L$. We define for each group the combined vectors and their centroid:

$$\mu_g = \frac{1}{N^2} \sum_{i=1}^{N} \sum_{j=1}^{N} \left( I_{g,i} \odot T_{g,j} \right). \tag{1}$$

where element-wise product is denoted by $\odot$.

To encourage the model to differentiate between distinct groups (a target group and its hard negative groups) of captions and images (*i.e.* support the outer group loss feature), we define for each image $I_{g,i}$, its positive texts $\{T_{g,j}\}_{j=1}^{N}$ (and vice versa for $T_{g,i}$). The two log-loss terms ensure that images are close to all texts in their group (and similarly for texts) relative to all negatives from other groups.

$$\mathcal{L}_{\text{pairwise}_{outer}} =$$
$$\frac{1}{2MN} \sum_{g=1}^{M} \sum_{i=1}^{N} \left\{ -\log \frac{\sum_{j=1}^{N} \exp\left(\frac{s(I_{g,i}, T_{g,j})}{\tau}\right)}{\sum_{g'=1}^{M} \sum_{j=1}^{N} \exp\left(\frac{s(I_{g,i}, T_{g',j})}{\tau}\right)} \right.$$
$$\left. -\log \frac{\sum_{j=1}^{N} \exp\left(\frac{s(I_{g,j}, T_{g,i})}{\tau}\right)}{\sum_{g'=1}^{M} \sum_{j=1}^{N} \exp\left(\frac{s(I_{g',j}, T_{g,i})}{\tau}\right)} \right\} \quad (2)$$

where $s(\cdot, \cdot)$ is a similarity function (*i.e.* cosine similarity), and $\tau > 0$ is a temperature parameter.

For the inner group loss feature (*i.e.* enhance the internal semantic coherence of each group) we treat each combined vector as a "query" that should match best with its own group's combined centroid, while the centroids from the other groups serve as negatives. In other words, we classify each combined vector into one of the $M$ groups based on its similarity to each group's centroid.

$$\mathcal{L}_{\text{pairwise}_{inner}} =$$
$$\frac{1}{MN^2} \sum_{g=1}^{M} \sum_{i=1}^{N} \sum_{j=1}^{N} \left[ -\log \frac{\exp\left(\frac{s(I_{g,i} \odot T_{g,j}, \mu_g)}{\tau'}\right)}{\sum_{g'=1}^{M} \exp\left(\frac{s(I_{g,i} \odot T_{g,j}, \mu_{g'})}{\tau'}\right)} \right] \quad (3)$$

The final pairwise loss is defined as

$$\mathcal{L}_{\text{pairwise}} = \alpha \mathcal{L}_{\text{pairwise}_{inner}} + (1 - \alpha) \mathcal{L}_{\text{pairwise}_{outer}} \quad (4)$$

where the hyperparameter $\alpha$ weight the two loss parts.

### 3.2.3. Centroid Oriented Contrastive Learning

This approach "compresses" each group into two summary vectors, one for images and one for texts. It simplifies the representation of the group and might be beneficial when the detailed pairwise interactions (captured by the element-wise products) are less important than the overall group summary. Similar to the Pairwise Oriented Contrastive Learning approach, for each group $g = 1, \ldots, M$ we have $N$ image–text pairs $\{(I_{g,i}, T_{g,i})\}_{i=1}^{N}$, $I_{g,i}, T_{g,i} \in \mathbb{R}^L$, and we define the per-group centroids (computed separately for images and texts) as

$$\mu_g^I = \frac{1}{N} \sum_{i=1}^{N} I_{g,i} \quad \text{and} \quad \mu_g^T = \frac{1}{N} \sum_{i=1}^{N} T_{g,i}. \quad (5)$$

The outer group loss term is defined as

$$\mathcal{L}_{\text{centroid}_{outer}} = \frac{1}{2M} \sum_{g=1}^{M} \left[ -\log \frac{\exp\left(s(\mu_g^I, \mu_g^T)/\tau\right)}{\sum_{g'=1}^{M} \exp\left(s(\mu_g^I, \mu_{g'}^T)/\tau\right)} \right.$$
$$\left. -\log \frac{\exp\left(s(\mu_g^I, \mu_g^T)/\tau\right)}{\sum_{g'=1}^{M} \exp\left(s(\mu_{g'}^I, \mu_g^T)/\tau\right)} \right] \quad (6)$$

where $s(\cdot, \cdot)$ is a similarity function (*i.e.* cosine similarity), and $\tau > 0$ is a temperature parameter.

The inner group loss term is defined as

$$\mathcal{L}_{\text{centroid}_{inner}} =$$
$$\frac{1}{2MN} \sum_{g=1}^{M} \sum_{i=1}^{N} \left[ -\log \frac{\exp\left(s(I_{g,i}, \mu_g^I)/\tau'\right)}{\sum_{g'=1}^{M} \exp\left(s(I_{g,i}, \mu_{g'}^I)/\tau'\right)} \right. \quad (7)$$
$$\left. -\log \frac{\exp\left(s(T_{g,i}, \mu_g^T)/\tau'\right)}{\sum_{g'=1}^{M} \exp\left(s(T_{g,i}, \mu_{g'}^T)/\tau'\right)} \right]$$

where $\tau' > 0$ is a temperature parameter.

Finally, $\alpha$ balances the two parts in the final loss

$$\mathcal{L}_{\text{centroid}} = \alpha \mathcal{L}_{\text{centroid}_{inner}} + (1 - \alpha) \mathcal{L}_{\text{centroid}_{outer}} \quad (8)$$

As can be seen, this approach reduces computational complexity because we're not considering all $N^2$ combinations within a group.

### 3.2.4. Textual Alignment Pretraining

To evaluate and reinforce the intuition on the hierarchical structure of abstraction within the textual latent space, we introduce an optional step of Textual Alignment Pretraining before training the final *CLEAR GLASS model* (similar intuition as [1]). In this phase, we use text-text (instead of image-text) contrastive learning to align captions with their textual concepts. This phase will allow us to assess the initial state of the textual embedding latent space, and to that extent, such an alignment contributes to the semantic abstraction understanding (see Sec. 4).

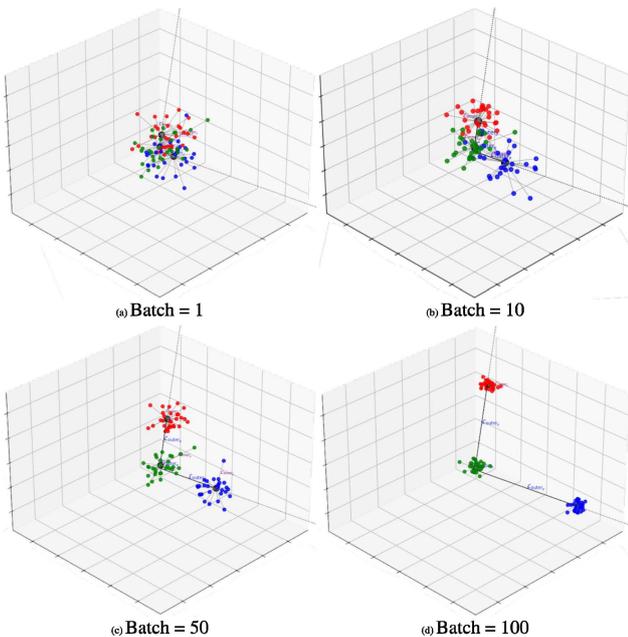

(a) Batch = 1  (b) Batch = 10
(c) Batch = 50  (d) Batch = 100

Figure 2. Illustration of the dual effect of Group Contrastive Learning loss: sharpening group distinctions and enhancing semantic tightness.

where each image-caption joint embedding vector is visualized as a dot, with each dot colored according to its group (*i.e.* concept).

| Hyper-Parameter | Search Range | Selected Value |
|---|---|---|
| Groups per batch | $[2\ldots10]$ | 2 |
| Pairs per group | $[2\ldots15]$ | 10 |
| Learning rate | $[1e-04\ldots1e-10]$ | $1e-08$ |
| $\tau$ (temprature) | $[0.1\ldots1]$ | 0.1 |
| $\alpha$ | $[0\ldots1]$ | 0.7 |

Table 2. Hyper Parameters Search

## 4. Experimental Setting & Evaluation

As detailed in Sec. 3.1, we generated the *MAGIC* dataset based on the *COCO* dataset (Tab. 1). *CLEAR GLASS* model was trained only on the *MAGIC* dataset *captions* and evaluated on the *MAGIC* on all levels, *HierarCaps* [1] on all levels, and *COCO* [25] on captions (as it contains only captions) datasets. As the model was trained only on the *captions* in *MAGIC*, we consider other levels (*i.e.* $l_1$ - $l_4$) as a benchmark (*i.e.* novel classes).

The model was trained using OpenCLIP [8] framework based on the ViT-B/32 base model (151M parameters). Using a Bayesian hyperparameter search [43] we identify optimal training parameters (Tab. 2) that maximize the average accuracy across varying settings of hyperparameters. During training, each dataset was randomly split into 80% training and 20% testing subsets, and each batch contained a group and its hard negative groups to increase the robustness of the model. Training sessions were done on a single GPU card (A5000) and were completed in ∼3 hours over five epochs, highlighting the efficiency of our method in enhancing abstract concept inference with minimal resources.

As illustrated in Fig. 4, and in greater detail in Tab. 3, our proposed model, *CLEAR GLASS* outperforms the baseline (*i.e. CLIP*) and the previous abstract concept inference model (*i.e. HierarCaps*). This is further supported by the hyperparameter search that found that the scaling hyperparameter is optimal at $\alpha = 0.7$, which indicates that both the inner and outer group loss terms are essential for inferring novel abstract concepts. Moreover, as can be expected, the higher the abstraction level - the lower the accuracy of all the models.

Although *CLEAR GLASS* was not trained on the abstract levels (*i.e.* $l_1$ - $l_4$), to ensure that the *CLEAR GLASS* does not overfit our dataset in any way and effectively enhances CLIP's accuracy on unseen data (*i.e.* generalization test), we evaluate the models on *HierarCaps dataset* [1], and *COCO* [25] datasets. The *HierarCaps dataset* was chosen for its unique focus on abstract concept inference, while *COCO* was selected to assess whether our method maintains accuracy on the standard contrastive learning task (image-pair matching). As demonstrated in Tab. 3, our method generalizes well on unseen data, improving accuracy even compared to models that were trained directly on these datasets.

Furthermore, the large accuracy gap between the accuracy achieved on *HierarCaps* and *COCO* datasets comparing the accuracy achieved on *MAGIC* highlights the challenging nature of abstract concept inference from images using vision-language models. These findings underscore the importance of the *MAGIC* dataset as a benchmark for future research aimed at incorporating abstract concept understanding into vision-language models.

### 4.1. Ablation Study of Group Contrastive Learning

To isolate the accuracy gain attributable solely to the Group Contrastive Learning methodology (Sec. 3.2.1) and to verify that *CLEAR GLASS*'s superior abstract concept inference is not merely a consequence of being trained on our dataset, we fine-tuned both the original *CLIP* model and *HierarCaps* on the same dataset (*i.e. MAGIC*) used for *CLEAR GLASS*. As shown in Tab. 3, *CLEAR GLASS* still outperforms the fine-tuned *CLIP* and *HierarCaps* models, indicating that the Group Contrastive Learning loss effectively leverages linguistic supervision to enhance semantic abstraction. Moreover, the fact that training on *MAGIC* did not improve the accuracy of the fine-tuned *CLIP* on the unseen attraction levels (*i.e.* $l_1$–$l_4$) further emphasizes the independent value of our approach as a whole.

| Model | Size | MAGIC | | | | | HierarCaps dataset | | | | COCO |
|---|---|---|---|---|---|---|---|---|---|---|---|
| | | captions | $l_1$ | $l_2$ | $l_3$ | $l_4$ | $l_1$ | $l_2$ | $l_3$ | $l_4$ | captions |
| CLIP [35] | 151M | 69.7 | 67.7 | 59.7 | 56.5 | 57.2 | 91.1 | 93.0 | 91.4 | 90.5 | 96.6 |
| CLIP$_{fine-tuned}$ | 151M | 74.2 | 68.7 | 59.5 | 55.9 | 57.4 | 90.4 | 92.9 | 91.1 | 90.3 | 94.1 |
| HierarCaps [1] | 151M | 71.0 | 68.0 | 59.2 | 56.0 | 56.6 | 91.5 | 92.7 | 89.7 | 84.4 | 96.0 |
| HierarCaps$_{fine-tuned}$ | 151M | 70.0 | 64.4 | 55.1 | 52.2 | 53.0 | 91.4 | 91.9 | 87.8 | 78.6 | 94.0 |
| CLEAR GLASS$_{centroid}$ | 151M | 74.1 | 70.2 | 61.3 | 57.3 | 58.9 | 92.4 | 93.6 | 92.6 | 91.4 | **96.8** |
| CLEAR GLASS$_{centroid_{pt}}$ | 151M | **77.4** | 73.6 | 61.2 | 58.2 | 57.6 | 91.5 | 92.9 | 87.0 | 83.4 | 88.1 |
| CLEAR GLASS$_{pairwise}$ | 151M | 74.8 | 71.1 | **61.7** | 57.3 | **59.1** | **92.8** | **93.9** | **93.7** | **93.6** | 96.2 |
| CLEAR GLASS$_{pairwise_{pt}}$ | 151M | 77.2 | **73.8** | 61.0 | **58.6** | 57.5 | 91.4 | 92.9 | 87.4 | 84.6 | 88.0 |
| Avg Accuracy Gain w.r.t: | | | | | | | | | | | |
| | MAGIC | 6.0% | 3.6% | 0.8% | 0.5% | 0.6% | 0.3% | 0.1% | -0.6% | -0.5% | -3.5% |
| | Loss Function | 6.5% | 7.4% | 5.0% | 4.8% | 3.9% | 1.0% | 0.7% | -2.5% | 2.6% | -3.0% |
| | Pretraining | 6.9% | 7.8% | 2.8% | 4.5% | 0.9% | -0.1% | -0.1% | -4.2% | -3.4% | -7.9% |
| | Best Model | 6.0% | 6.8% | 3.5% | 4.2% | 3.8% | 1.5% | 1.1% | 4.6% | 8.6% | 3.7% |

Table 3. Accuracy (text to image R@1) of *CLEAR GLASS* comparing to CLIP and other SOTA models.
where *CLEAR GLASS$_{pairwise}$* was trained using the pairwise loss (Eq. (4)), *CLEAR GLASS$_{centroid}$* was trained using the centroid loss (Eq. (8)), *CLEAR GLASS$_{pairwise_{pt}}$* was pretrained for textual alignment (Sec. 3.2.4) and trained using the pairwise loss, *CLEAR GLASS$_{centroid_{pt}}$* was pretrained for textual alignment and trained using the centroid loss, *CLIP$_{fine-tuned}$* and *HierarCaps$_{fine-tuned}$* were trained on the *MAGIC* dataset, abstraction level represents the level of abstraction within each dataset where *captions* represents the explicit captions (*i.e.* no abstraction), $l_1$ represents the first level of abstraction, $l_2$ represents the second level of abstraction, etc.

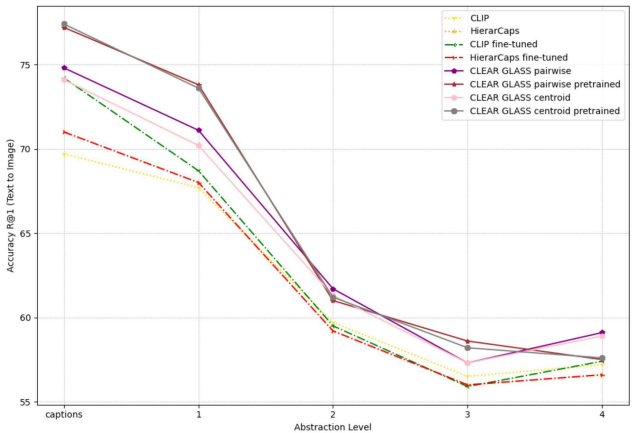

Figure 4. The change in accuracy with respect to linguistic abstraction levels in the *MAGIC* dataset

When comparing the Centroid Oriented Contrastive Learning (Sec. 3.2.3) and the Pairwise Oriented Contrastive Learning (Sec. 3.2.2) strategies on both *MAGIC* and *HierarCaps* (Tab. 3), it is evident that the pairwise-oriented loss better captures the nuances of abstract concepts. This suggests that preserving fine-grained alignment information—since the pairwise approach operates on individual image-text pairs—enables the model to discern subtle differences necessary for inferring abstract concepts.

To quantify the average accuracy gain of each component, we compare: (i) models trained without vs. with the *MAGIC* dataset, (ii) models without vs. with a Group Contrastive Learning loss (pairwise or centroid), (iii) models without vs. with pretraining, and (iv) the best model. As demonstrated in Tab. 3, each component contributes to the final accuracy. As expected, the impact diminishes at higher abstraction levels. However, as suggested by other findings, the influence of the dataset decreases far more rapidly than the other components. Furthermore, the Group Contrastive Learning loss functions contribute the most to the accuracy and allow generalization, while the pertaining process clearly overfit the model on our dataset and even reduces accuracy (on average) on other datasets.

## 5. Conclusion and Discussion

This paper introduces a novel group-based approach to contrastive learning (Group Contrastive Learning) in order to induce visual language models (VLMs) to recognize higher-level conceptual information in images. Recognizing such abstract, hierarchical concepts remains a challenge, as existing datasets and methodologies lack the capacity to capture the semantic depth required for this task. Since there is no ready-made dataset for this purpose, we created a hierarchical dataset (*MAGIC*) consisting of a tree of concepts in which images (and their corresponding captions) are organized into multiple levels of semantic abstraction. Using this dataset and the Group Contrastive Learning method, we developed a model (*CLEAR GLASS*) that shows a greater capacity to recognize higher-level conceptual information in images. In developing this model we use a novel loss function which is the scaled summation of the outer group loss and the inner group loss. The outer group loss is intended to encourage the model to distinguish between a true group of captions for a group of images and a semantically closely related group of captions (hard negatives) which are not the correct captions for those images. The inner group loss is intended to enhance the internal semantic tightness of each group. We evaluate the model in terms of whether or not the model recognizes in the hierarchical tree of concepts the parent concepts of the true group of captions for a group of images. Since these parent concepts are never exposed to the model during training, our evaluation strategy measures the degree to which the model exhibits emergent learning of concepts. Our experimental results demonstrate that, while our approach is promising, the task remains challenging - especially at higher abstraction levels, with our best model achieving only 59.1% accuracy at level 4 abstraction. Notably, our method outperforms both the baseline model, *CLIP*, and the current SOTA abstract concept inference model, *HierarCaps*, while also generalizing effectively to unseen datasets. Furthermore, our analysis reveals that the Group Contrastive Learning loss functions have the most significant impact on accuracy, whereas the dataset primarily enhances performance at lower abstraction levels. In contrast, the pretraining approach tends to overfit the model, thereby impeding generalization.

Advancing abstract concept inference in images enhances various tasks. For example, image search can retrieve results for high-level terms like *celebrations* even when captions lack that word. In robotics, it enables natural commands such as "Clean all the furniture but avoid electronic equipment," even if the training data lacks explicit references to furniture or electronic devices. Similarly, in visual question answering, systems can address queries like "What clothing is appropriate for sports events?" using images labeled with events such as "a baseball game at Yankee Stadium" or "a basketball match at my school."

Looking ahead, we plan to address these challenges by extending our method with abstract concept anchoring via linguistic supervision, exploring architectural changes (*e.g.* mixed experts) to enhance multi-level abstraction, and adding 'abstract adapters' to boost model composability. These advancements aim to push VLM boundaries and drive positive impacts across fields.

## 6. Acknowledgement

This research work was supported by research grants from the Data Science Institute of Brown University to Shekhar Pradhan and to UTRA awards from Brown University to Muhiim Ali.